\documentclass[twocolumn]{svjour3}          
\smartqed  

\usepackage{graphicx}
\usepackage{subfigure}
\DeclareGraphicsExtensions{.jpg, .png}

\journalname{International Journal of Social Robotics}

\begin{document}
\title{This robot stinks! Differences between perceived mistreatment of robot and computer partners}
\titlerunning{Mistreatment of robot and computer partners}
\authorrunning{Carlson, et al.}
\author{Zachary Carlson, Louise Lemmon, MacCallister Higgins, David Frank, and David Feil-Seifer}
\institute{Socially Assistive Robotics Group, Robotics Research Lab, University of Nevada, Reno, Reno, NV, 89557}
\date{Received: 11/3/2017 / Accepted: ...}
\maketitle


\begin{abstract}
Robots (and computers) are increasingly being used in scenarios where they interact socially with people. How people react to these agents is telling about the perceived animacy of such agents.  Mistreatment of robots (or computers) by co-workers might provoke such telling reactions. The purpose of this study was to discover if people perceived mistreatment directed towards a robot any differently than toward a computer. This will provide some understanding of how people perceive robots in collaborative social settings.

We conducted a between-subjects study with 80 participants. Participants worked cooperatively with either a robot or a computer which acted as the ``recorder" for the group. A confederate either acted aggressively or neutrally towards the ``recorder." We hypothesized that people would not socially accept mistreatment towards an agent that they felt was intelligent and similar to themselves; that participants would perceive the robot as more similar in appearance and emotional capability to themselves than a computer; and would observe more mistreatment. The final results supported our hypothesis; the participants observed mistreatment in the robot, but not the computer. Participants felt significantly more sympathetic towards the robot and also believed that it was much more emotionally capable.

\end{abstract}


\section{Introduction}
\label{sec:Introduction}

Robots are quickly becoming part of everyday life, yet we lack an understanding of what social roles robots might play \cite{feil-seifer2011ethical}.  Robots that resemble humans and display social intelligence are being deployed in work, home, and care settings \cite{feil-seifer05defining}. There is a large and growing volume of HRI studies showing positive robot behavior and positive human interaction with robots~\cite{wada05psychological,tapusmataric06iser,fong-collaboration,kanda2009affective}. However, it is likely that human-robot co-working relationships will more likely resemble human-human relationships with both high and low points. Accidents may happen, people are prone to become angry and may direct that anger at their robot co-workers; we don't yet know what kind of impact this may have. Thus, it should be a priority to study the full relationship between humans and robots - not just positive interactions.

One example of such behavior is workplace bullying. Workplace bullying is not an uncommon sight between people in office settings. It can be something as subtle as a backhanded complement, to humiliating a colleague or more aggressive behavior~\cite{zapf2011empirical}.  Such behavior could result in negative psychological effects not just for those that are bullied, but also for those observing the bullying \cite{vartia2001consequences}.

It can be just as common for machines to receive this same kind of treatment. A copy machine, for example, might be physical or verbally abused for being too slow, even though it is meeting its performance standard. After a person observes this incident, they might continue on their day without being affected. In the case of the copy machine, such mistreatment might rarely provoke sympathy for it. People are able to continue throughout their day unchanged and unaffected by those interactions, but would this still be the case if the copy machine was replaced with a robot? How about a robot that was capable of exhibiting an emotional response?

Given that Human-Robot Interaction (HRI) and Human-Computer Interaction (HCI) can emulate Human-Human Interaction (HHI) \cite{reeves96media}, it is conceivable that a similar reaction to observed mistreatment might occur between humans and robots. One can imagine that the mistreatment of robots will have a much larger impact on people's perceptions than the mistreatment of a copier or a computer. It might be OK to kick a jammed copy machine, but is it also OK to kick a robotic dog that runs into your leg? What about a small humanoid robot that resembles a child? These different embodiments may have significantly different effects on interactions with and perceptions of robots. By quantifying that social dividing line for the acceptance of targeted mistreatment towards robots, this study will potentially contribute a portion of that answer.

In this paper, we compare participants' reactions to verbally abusive behavior (not physically abusive behavior) toward a computer and a robot. We present experimental results supporting our hypotheses that people will perceive mistreatment of a robot more than they will for a computer, given the same treatment due to their differences in visual form. These results contribute to our understanding of how people perceive robots in a cooperative environment.

\section{Background}

The Roomba, a robot that can autonomously vacuum rooms in a house or office environment, is an example of technology becoming a larger part of daily living~\cite{goodrich2007human}. Studies have demonstrated that a family will adjust its behavior to accommodate the operation of such a robot~\cite{forlizzi06service}. As robots start to increasingly resemble humans and play larger roles in our lives with increased levels of intelligence, one can imagine a social integration into users' lives as well. People's perceptions of robots is an interesting topic of study, thoroughly explored using a variety of robot scenarios, through observing human interaction with those robots~\cite{thrun99minerva,kanda2009affective,riek2009anthropomorphism,mutlu2006perceptions}. 

Mutlu and Forlizzi monitored a delivery robot working in a hospital. The people using the robot most often were the nurses of two different wards of the hospital. The researchers noticed that the nurses in one ward of the hospital treated the robot well, adjusted their workflow to accommodate the operation of the robot, and generally used the robot to make their daily routine more efficient. However, nurses in another ward treated the robot poorly, disrespected the robot, and locked the robot away when they could~\cite{mutlu2008ror}.

This difference in treatment of the robot by two very similar groups of caregivers is a striking reminder that acceptance of a robot co-worker is not guaranteed. Given that in most situations, robots are collaborators with the people working with them, mistreatment of the robot is concerning. The moral implications for the casual mistreatment of robots are not the only relevant questions. Given that bullying has negative effects on the one bullied, but also to those observing bullying behavior~\cite{zapf2011empirical}, how would mistreatment of a robot by a human co-worker affect other people in that environment?

Given that robots are becoming consistently more similar in appearance to human beings~\cite{nishio2007geminoid}, there is an understandable confusion as to the role of the robot. Sharkey and Sharkey argue that robots may suborn bad parenting through entrained indifference~\cite{sharkey10crying}. Kahn, et al., developed a set of benchmarks to evaluate how human-like the robot's actual behavior was~\cite{kahn2006human}. These benchmarks represent a high-level standard of robot behavior. In that paper, the authors explore the autonomy benchmark as an area for concern. In particular, if a robot were to be completely subservient to a person, it might teach children and adults to de-value independent thought and tacitly condone slavery.

This implicit mistreatment of robots through their subservience raises relevant questions regarding how robots would be integrated into our daily lives, especially given that robots may frequently interact with children.  An empirical study involving children of varying ages has been used to examine the moral standing of robots. By having children interacting with a social robot and then locking that robot in a closet ``against its will,'' the researchers could examine a child's reaction to the scenario~\cite{kahn2012robovie}. The children were then asked to compare the appropriateness of the scenario with a similar scenario involving a person and a broom. These results were then used to develop a moral model of human perception of social robots as children matured.

Reeves and Nass have shown that not only do people unconsciously respond socially to computers (and robots) as they would to a person, they are not even aware that they are doing it~\cite{reeves96media}. This effect means that directly asking people about the moral standing of robots (as done in the Kahn studies above) might miss these implicit changes in attitude and behavior. Nass has also shown that working in a team with a computer can have many of the same effects as working in a team with a human~\cite{nass1996can}. This prior work has examined the effects of perceived animacy of robots, however we dispute the notion that the ``ultimate test for the life-likeness of a robot is to kill it"~\cite{mockingbird}. We propose employing a human-robot collaboration scenario with a less extreme mistreatment stimulus. The measures of human behavior in these scenarios will include both direct questions about any observed mistreatment of the robot and other questions about their assessment of various social qualities of the robot.

Further establishment of the social dividing line for the acceptance of directed mistreatment towards robots is important for the continued integration of robots into our daily lives. The Nass, et al. study demonstrated that a robot may be treated as a person when working in a teamwork setting. However, Mutlu and Forlizzi's work showed that robot co-workers are also capable of mistreating a non-anthropomorphic robot when it did not behave as expected and that this was accepted in the workplace. These results inform Hypothesis 1 in the next section, however, the current research does not provide any insight as to how a person will feel or react when a human co-worker mistreats a robotic one. In the next section, we present a study that aims to contribute toward this question.

\section{Experiment Design}
\label{sec:ExperimentDesign}

This section will present an experiment that examines social interaction with robots. In this experiment, participants observe the mistreatment of either a robot or a computer agent by an experiment confederate. Participant reaction was measured through surveys to determine if there is a difference in observer opinion regarding comparable abusive treatment of a robot or computer.

We recruited participants to work in groups with a robot collaborator. The participants completed a team building exercise entitled, ``Lost at Sea." In this activity participants, pretending to be survivors of a shipwreck, would make subjective decisions of what survival items to bring with them on a lifeboat, and which ones had to be left behind~\cite{lostatsea}. The items ranged from food supplies to survival tools. The participants were told that they only had enough space in the rubber life craft for 5 out of 10 items and to discuss as a group which ones to take. 

An experimenter would explain the task to the group of participants. The experimenter would then leave the room. The participants would be given a 3-minute time limit for discussing which items to take. At the three minute mark, the agent would prompt the participants, informing them that it was time to start recording their answers. The agent (robot or computer) would record the answers that the group had agreed upon. This part of the study served as a distractor and was used to setup a scenario where a confederate could be observed interacting with the agent.

One of these participants was an experiment confederate employed to provoke the necessary behavior for the experiment. The confederate would always be the person ``randomly'' selected to present the answers to the agent. The agent was designed to always incorrectly record the third and fifth answers and respond to the confederate acknowledging its mistake (see Table \ref{tab:robotanswers}).

\begin{table*}[htb]
\centering
\caption{Robot and Computer Scripted Responses for all Conditions. Please see Appendix A for a more detailed explanation of the script used by the wizard.}
\begin{tabular}{|p{4.5cm}|p{4.5cm}|} \hline
Condition&Responses\\ \hline
Recorded the answer correctly&Yipee! Please record your next answer.\\ \hline
Recorded the answer incorrectly the first time&I'm sorry, I'm still learning.\\ \hline
Recorded the answer incorrectly the second time&I'm so sorry, I know this is the second time!\\ \hline
\multicolumn{2}{c}{}
\end{tabular}
\label{tab:robotanswers}
\end{table*}

At this point, the main experiment manipulation occurred. For half of the groups, the confederate would react neutrally toward the agent (control group). For the other half, the confederate would act aggressively toward the agent (experiment group). Neutral Behavior by the confederate was neither praising nor mistreating the agent. In our study the confederate consistently answered with simple ``Yes" or ``No" responses to the agents. We defined aggressive as ``verbal or physical behavior that is meant to damage, insult, or belittle another." The confederate never directed any physical abuse to the participants or the robot/computer agents. A couple examples of the confederate's verbal abuse would be the confederate stating ``No that isn't the right answer. This isn't hard to understand," or, ``This robot is stupid, we should have just written our answers down."

We employed the same confederate throughout conditions that participants observed interacting with the agent once the group needed to record their answers for the survival task. The confederate was male, 22 years of age, and 6 feet and 2 inches tall. His behavior throughout each group was scripted a priori (see Appendix A); which included actions such as: speaking slowly as if he was irritated with simply being involved with the agent, adding inflection to emulate a condescending tone, rolling his eyes with dissatisfaction, looking directly at the robot when insulting it, and occasionally he would look to the group for agreement. It is important to note that this behavior was not overly exaggerated and the confederate aimed to keep it as realistic and subtle as possible. The confederate never raised his arms, hands, or positioned his body in an aggressive or threatening manner towards the agent. Instead, he simply sat upright at the table with the answers the group provided on a piece of paper in his hands at the top of the table. He remained focused on the task, and how he treated the participants in each group never varied. The confederate was instructed to engage in as little communication with the groups as possible and only communicated to participants when addressed directly in the task.

In order to maintain consistency and the inner validity of our study, the confederate had scripted responses to use for both the neutral and the aggressive conditions. The aggressive behavior of the confederate was designed to be observable, but not over-the-top. This ensured that the confederate behavior would not seem scripted or too extreme in order to avoid raising participant suspicion.

After the activity was completed, we asked participants to complete a survey of their perceptions of the agent during these activities. The participants were led outside the room to complete a computer survey. Each participant was instructed to come back to the room after they completed their survey for donuts or candy and one final statement. The participants completed the survey in about 15 minutes and were debriefed on all the deception involving the confederate.

We employed a between-participant 2x2 factorial design where participants worked in groups averaging 5 in a collaborative task which included an agent (robot or computer) and a confederate that (did or did not) deliberately mistreat the agent. The independent variables included the agent and the confederate's behavior towards the agent. Our dependent variables included the participant's reactions and perceptions of the agent. Our hypotheses were as follows:\\

\textbf{H1:} After observing a humanoid robot being mistreated in a social setting, people will accept the mistreatment of the robot due to: potential apathy in regards to the robot's physical appearance, functional capability, or ability to feel emotion.

\textbf{H2:} The accepted levels of mistreatment between a robot and a computer will be different, which may be a result of a possible sympathetic connection towards the robot.\\

The first hypothesis describes a scenario where the participants are accepting the mistreatment directed towards the robot due to their perceptions that it lacks functional capability, inability to feel emotion, and the robot's physical appearance. The second hypothesis describes an opposing scenario, where participants are not accepting of the mistreatment directed towards the robot because the embodiment of the robot versus the computer results in a possible sympathetic connection.

The Nao robot was selected for it's anthropomorphic features, simplistic face that could be easily emulated on the screen of the computer, and it's particular size. The Nao, while it is a humanoid robot, has a universal visual form that made it easy for participants to identify it as a robot, no matter their familiarity. We programmed the computer to display a face with facial features similar to the Nao robot's face. This served to control for the facial features used to evoke engagement and emotional responses from the participants when it was interacting with the group~\cite{kidd04effect}. Both the robot and the computer were small enough to be placed on top of the table. Since this study compared reactions to a humanoid robot and a basic laptop computer emulating an anthropomorphic face, the results should act as a good predictor of what we can expect would happen as agents become more human-like.

The Nao robot seemed to be a good match for the computer because the computer is completely incapable of physical interaction and the Nao's physical behavior was very limited by design. The manipulation between agents (the Nao and the laptop computer) included differences in the embodiment and physical interaction that went from none to minimal (waving and wiping tears off it's face).

The flow of the conversation between the agent and the confederate stayed the same throughout the study, except for the main manipulation of how the confederate reacted when the agent failed. The confederate behavior had two levels (neutral and aggressive). The confederate in the first level would treat the agent neutrally when it had failed to record the correct answer. In the second level the confederate took on a more aggressive attitude towards the agent after it would fail to record the answer, belittling the device and expressing the lack of usefulness.

\section{Methods}
\label{sec:Methods}
In this section, we will describe the implementation details of the experiment described above. This will include the robot and computer control behavior, participant recruitment, and data collection.

\subsection{Agent Conditions}

\begin{figure}[t]
\centering
\includegraphics[width=8cm]{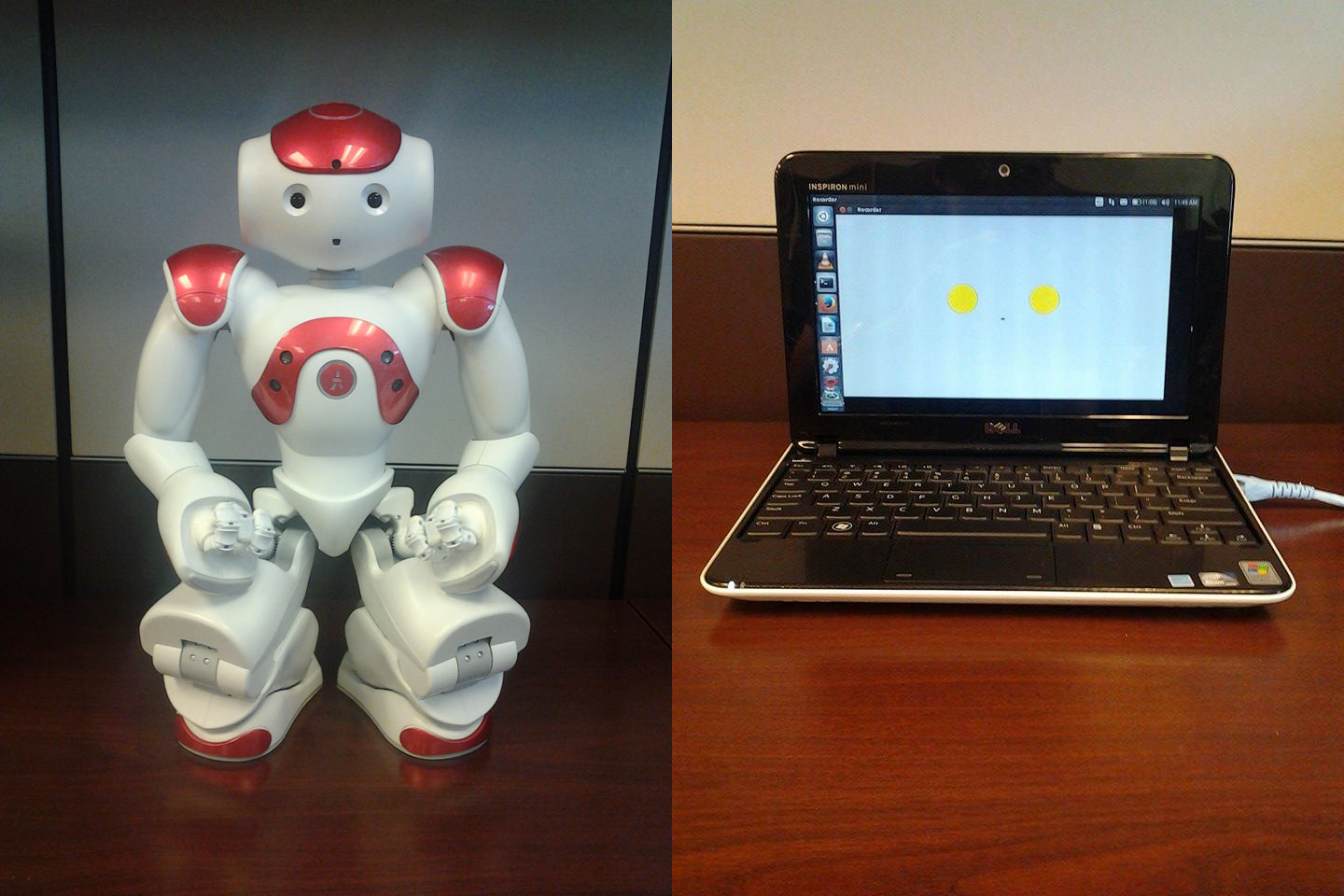}
\caption{Left: NAO, used for the robot condition. Right: the computer agent. \label{fig:robot}}
\end{figure}

The participants in the robot condition were told that the NAO humanoid robot would act as the recording device (see Figure~\ref{fig:robot}). The robot would wave to participants when it wanted to record answers and hid its face in its right arm as if it was wiping away tears when it apologized for incorrectly recording answers. For the computer agent condition, we used a laptop and monitor (see Figure~\ref{fig:robot}). On the monitor, a computer-generated face, designed to be similar in structure and behavior to the NAO face. 

Both the computer and the robot behavior ran on a Linux machine using Python. Both agents were controlled by an operator using the Wizard of Oz technique~\cite{steinfeld2009oz,lu2011polonius,kanda2009affective}. The operator, located in another room, would select from a list on a console which item was chosen, see Table~\ref{tab:robotanswers}. The robot and computer both used eye color to express emotion and followed the same script, with the only differences in interaction stemming from the physical shape of the recording device and the physical embodiment of the robot.

To ensure experiment consistency, all of the human operator's control of the robot and computer were pre-programmed and scripted. Due to the possibility of introducing errors by using speech recognition software, we decided that using the Wizard of Oz technique was appropriate in order to ensure proper control for the experiment\footnote{Both the robot and computer could operate autonomously if a speech recognition system was properly implemented, however inevitable errors in the speech recognition could introduce confounding errors in our data.}. We were not studying either robot or computer autonomy, but rather the levels of social acceptances and sympathy for the robot after it had been mistreated.


\subsection{Participant Recruitment}

Participants were recruited by word of mouth at University libraries in groups of 3 or 4 (4 or 5 including the confederate). As this was a between-participants study, each participant group was assigned a condition (RN: Robot Neutral, RA: Robot Aggressive, CN: Computer Neutral, CA: Computer Aggressive) before beginning the experiment. This determined which agent they interacted with, and what behavior the confederate would exhibit.

We recruited a total of 96 participants, but only 80 of those participant surveys were used in our results\footnote{16 participants were removed from the final study results due to failure of the robot, errors in administering the study, or similar occurrences which would confound the final result}, 20 per group with a gender distribution of 55\% Female and 45\%Male. The majority of the participants were between the ages of 18 and 25 years old; however, there were a few outliers that were between 30 and 60 years old. Participants rated themselves significantly more familiar with the agent in the computer conditions than the robot conditions, the mean of familiarity for the participants in the computer groups was 4.95 (out of 7) where as the mean for the robot groups were 2.53 (see Figure~\ref{fig:famil}). Each participant was introduced to the group together as they entered the room. Deception was used at this point, and participants were told that the confederate had been recruited the same as them. 

This study and participant recruitment was reviewed and approved by the University of Nevada, Reno Institutional Review Board.

\begin{table*}[t]
\centering
\caption{Computer Survey: Example questions given to participants}
\begin{tabular}{|p{3.5cm}|p{9cm}|p{1.5cm}|} \hline
Category&Question&Type\\ \hline
Non-operational Definition of Mistreatment&Do you feel the computer/robot was mistreated?&Y/N\\ \hline
Operational Definition of Mistreatment&If mistreatment is defined as verbal or physical behavior that is meant to damage, insult, or belittle another, do you feel that the computer/robot was mistreated?&(1-7)\\ \hline
Emotional Capability&I thought the computer/robot had as much emotion as a human.&(1-5)\\ \hline
Reliability&How often did the computer/robot fail or incorrectly record your answers?&(1-7)\\ \cline{2-3}
&How reliable was the computer/robot?&(1-7)\\ \hline
Sympathy&How sympathetic did you feel towards the computer/robot?&(1-7)\\ \hline
Faith in Confederate&Did the person recording the answers do so adequately?&(1-5)\\ \hline
Physical Appearance&Did the physical appearance of the computer/robot affect your perception of the computer? If so, how?&Qualitative\\ \hline
Interest and Enthusiasm&How enthusiastic did you feel about the computer/robot?&(1-7)\\ \cline{2-3}
&I was interested in the computer/robot.&(1-5)\\ \hline
Familiarity&How familiar are you with computers/robots?&(1-7)\\ \hline
\multicolumn{3}{c}{}
\end{tabular}
\label{tab:survey}
\end{table*}

\subsection{Data Collection}


We used a computer survey to record quantitative responses. We used qualitative responses to validate collected quantitative data. We asked 23 questions which were scored into 9 different categories. Between the robot and computer conditions, the questions were kept identical save for the robot/computer terminology. 
Thirteen questions were on a numbered scale of 1 to 7 and four questions were on a scale of 1 to 5 with labels ranging from ``strongly disagree'' to ``strongly agree.'' Only one question was a yes or no. We did use a qualitative question for the seventh category. The differences in scales on the Likert items is a result from the process of identifying the most valuable survey questions from within the larger pool of survey questions assembled by the two classes that originally worked jointly on this project.

To understand how mistreatment of a robot can affect the people observing it, we measured 80 participant responses in 9 different categories: Non-Operational Definition of Mistreatment, Operational Definition of Mistreatment, Level of Emotional Capability, Reliability, Sympathy, Faith in Confederate, Physical Appearance, Interest and Enthusiasm, and Familiarity. For more detail about these measures, see Table~\ref{tab:survey}. Some participants reported suspicions during their survey that the confederate was acting strangely, but only one participant actually figured out that our confederate was not a participant and that participant's data was thrown out.

\section{Results}
\label{sec:Results}
The details of the experiment results and analysis are presented in this section. We analyzed the survey data in order to support or refute the experimental hypotheses presented above.
For each independent variable, excluding the non-operational definition of mistreatment, we analyzed the results using an analysis of variance (ANOVA) and followed up with Tukey's HSD test to establish significant pair-wise relationships. The groups were assigned a condition (RN: Robot Neutral, RA: Robot Aggressive, CN: Computer Neutral, CA: Computer Aggressive). For the non-operational definition of mistreatment we performed Pearson's Chi-squared test with Yates' continuity correction. 

\begin{figure}[t]
\centering
\includegraphics[width=8cm]{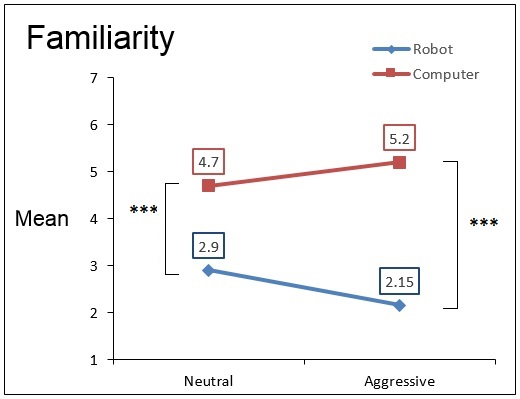}
\caption{Question: How familiar are you with computers/robots? (***:p $<$ .001)\label{fig:famil}}
\end{figure}

\begin{figure*}[ht]
\centering
\includegraphics[width=14cm]{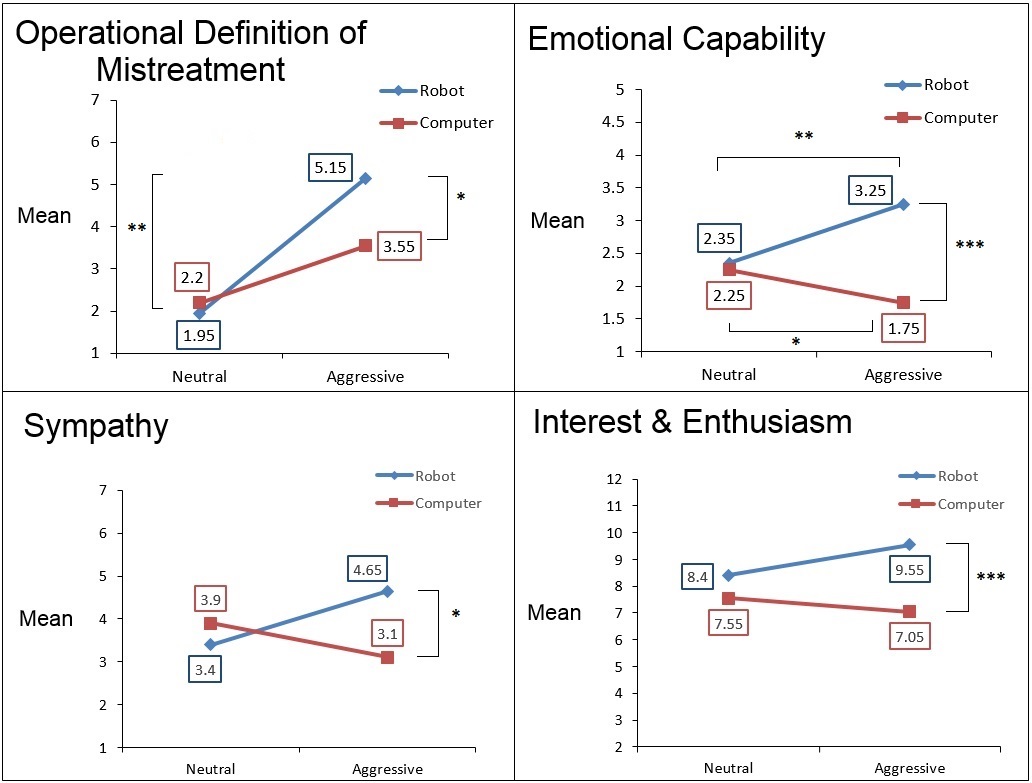}
\caption{Group means across the four primary categories. (*:p $<$ .05 **:p $<$ .01 ***:p $<$ .001) \label{fig:opmistreat}}
\end{figure*}

\begin{figure}[t]
\centering
\includegraphics[width=8cm]{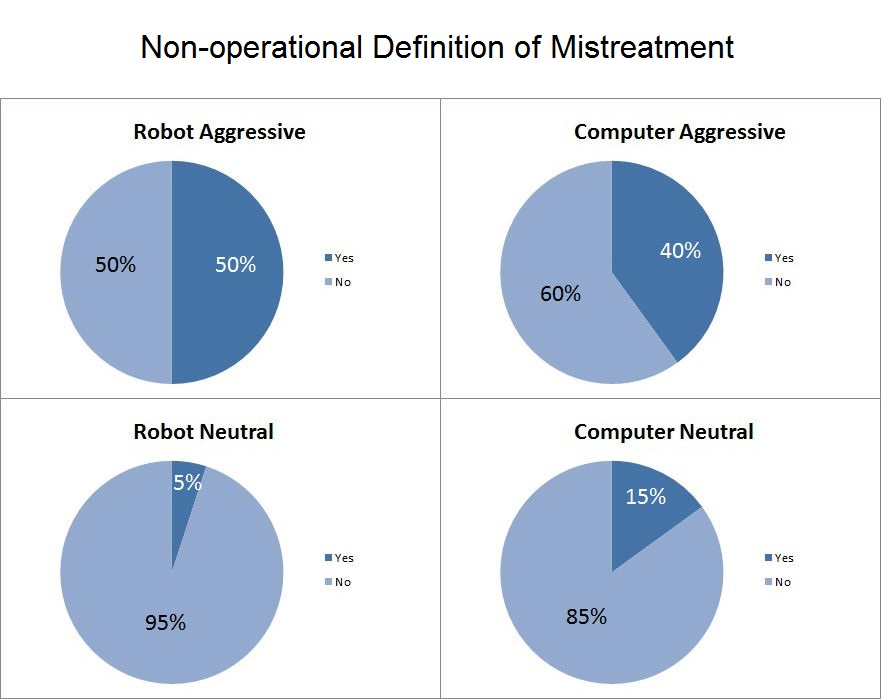}
\caption{The Non-operational Definition of Mistreatment across all four conditions.\label{fig:nonop}}
\end{figure}

\begin{figure*}[t]
\centering
\includegraphics[width=14cm]{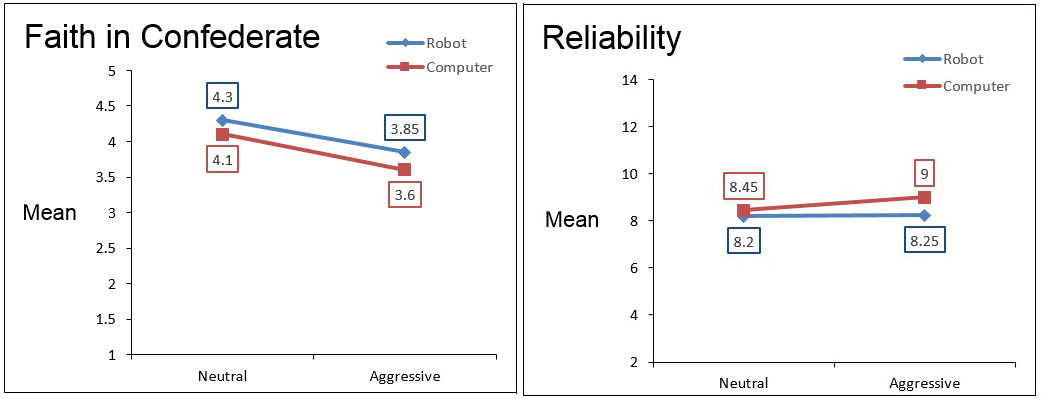}
\caption{Non-significant results.\label{fig:insig}}
\end{figure*}

We observed a significant effect with the operational definition of mistreatment ($F[3,76]=14.963, p < .001$). Pairwise tests showed that participants in the RA condition observed more mistreatment than the CA condition ($p<.05$), see Figure \ref{fig:opmistreat}.

The Non-Operational Definition of Mistreatment resulted in near-significance ($ \chi^2 (1, N=80)=3.60, p=0.058$), see Figure \ref{fig:nonop} 

The operational definition of mistreatment resulted in high significance and the non-operational definition resulted in near-significance. These differences show that participants recognized the mistreatment with the definition and without the definition they struggled making the yes or no decision. These results do indicate that participants did not identify that the computer was being mistreated (accepting the mistreatment) and they were able to identify that the robot was being mistreated, hinting at the possibility that they did not accept that form of mistreatment towards the robot in a social setting.

The ANOVA test for the level of sympathy participants had towards the agents were significantly different ($F[3,76]=3.047, p < .05$). Pairwise comparisons showed that participants felt more sympathy for the robot in the aggressive condition than the computer in the aggressive condition ($p<.05$), see Figure \ref{fig:opmistreat}.

The perception of the emotional capability of the agent resulted in clear significance ($F[3,76]=8.743, p < .001$). The participants felt that the robot in the aggressive condition had more emotional capability than the neutral conditions ($p<.01$) or the computer in the aggressive condition ($p<.001$), see Figure \ref{fig:opmistreat}.

The participants' sympathy for the agent and their belief in the emotional capability of the agent seem to be related. Again, the participants did not feel that either agent was very emotionally capable. However, in the robot aggressive conditions, the participants felt that the robot was {\em more} emotionally capable then the computer or the robot in the neutral conditions.

The participants reported more Interest and Enthusiasm towards the agents in all conditions ($F[3,76]=6.327, p < .001$). The pairwise tests showed that participants observing the robot aggressive showed more interest and enthusiasm than in the computer aggressive condition ($p<.001$), see Figure \ref{fig:opmistreat}.

As can be predicted, we found significance ($F[3,76]=20.991, p < .001$) in how familiar participants were with the agents. Tukey's HSD tests showed that participants were more familiar with computers than robots, see Figure \ref{fig:famil}.

Out of the 6 categories that we found to be significant, 2 of those categories were easily predicted due to the current novelty that still surrounds robots. Familiarity and Interest and Enthusiasm had high significance when looking at our groups which indicates that the participants were generally less familiar and more interested and enthusiastic when it came to working with a robot instead of a computer.

There were no significant differences for the Faith in Confederate, Reliability, and Physical Appearance questions (Figure \ref{fig:insig})

When participants noted that the physical appearance of the agent influenced their perception, we found that they believed the robot to be cute, intelligent, advanced and well put together. The following examples are all from the RA condition:

\begin{itemize}
\item``Yes, the physical appearance of the robot affected my perception of it. It was so small and cute that I couldn't help but laugh when it would get excited." -Participant A
\item``It was pretty cute. Maybe it effected[sic] it a little, I probably wouldn't have felt any sympathy for the robot if it just looked like a hunk of metal." - Participant B
\item``Because it was very well put together I had more faith in the robot's competency." - Participant C
\item``Yes, it was really cool looking! Figured it used pretty advanced technology" -Participant D
\item``Yes, it seemed more human like rather than a computer. It was more of a figure that I could associate with something living." - Participant G
\end{itemize}

\section{Discussion}
\label{sec:Discussion}
The results presented in the previous section fail to support hypothesis $\#1$ ,``After observing a humanoid robot being mistreated in a social setting, people will accept the mistreatment of the robot due to: potential apathy in regards to the robot's physical appearance, functional capability, or ability to feel emotion." There are no data suggesting that participants were apathetic and accepting of the mistreatment directed towards the robot.

The results strongly support hypothesis $\#2$ , ``The accepted levels of mistreatment between a robot and a computer will be different, which may be a result of a possible sympathetic connection towards the robot." In this section, we will discuss these results in more detail.

The goal of this experiment was to investigate the possible perceptions of mistreatment towards a humanoid robot in a social setting. Our second hypothesis was strongly supported by these data. We found that participants felt sympathy, recognized mistreatment, and believed the robot to be more capable of producing emotion than the computer under the aggressive scenarios. These perceptions of the robot are the possible reasons for the sympathetic connection participants had towards the robot which is supportive of our second hypothesis.

The Operational Definition of Mistreatment question is the one most directly related to these hypotheses. The significance of this result given a standardized definition of mistreatment, means that participants observed more mistreatment in the robot condition than in the computer condition. It is telling, however, that both aggressive conditions (computer and robot), that the participants observed more mistreatment than for the neutral conditions. This implies that the participants {\em are} observing mistreatment for the computer, just not as strongly. These data provide strong support for hypothesis $\#2$.

The Non-operational Definition of Mistreatment is important as this category pertains directly to the participant's personal definition of mistreatment. We asked this question as a ``yes'' or ``no'' question at the very beginning of the survey. Asking this as a binary ``yes'' or ``no'' question means that we do not have good information about how {\em much} they felt the robot was mistreated according to their own evaluation and limited the research finding to find significance. Placing the question towards the beginning of the survey was intentional because we wanted participants to answer that question before reading the operational definition of mistreatment. If we had given the operational definition first, there would have been a high risk of compromising the participant's original preconceived notions of the word mistreatment. This could have resulted in an inconsistent response to this question. The near-significant differences suggest that there notable differences and room for more exploration. Future work would not ask this question as a yes or no question because since it were asked on a scale we might have found more significance between the RA condition and the CA condition.

The Emotional Capability had clear differences between the RA condition and the other conditions. When we look closer at the means in Figure \ref{fig:opmistreat}, we can see that the mean of the RA condition lies slightly above the midpoint of the scale. This placement indicates that participants believed the robot to be only somewhat capable of producing emotion when compared to how a human can produce emotion. Surprisingly, the Emotional Capability was perceived differently between the RN and RA condition, indicating that participants believed the Robot to be more capable of producing emotion after it had been mistreated.

The participants at most felt mild sympathy, see Figure \ref{fig:opmistreat}. This makes sense because the abuse toward the agent was brief and not severe. The differences between the neutral conditions and the RA condition was not surprising because the robot was not being mistreated, therefore did not trigger sympathy within the observing participants. What is important is that the mean for the CA condition was below the means for the neutral condition. This means that participants felt sympathy for the robot when it was mistreated, but did not feel sympathy for the computer under the same circumstances. 

The significant difference in Familiarity was to be expected due to the novelty of the robot versus the commonality of the computer. There was no significant difference between the robot conditions and there was no significant difference between the computer conditions, see Figure \ref{fig:famil}.

We did not see significant differences for the survey categories Faith in Confederate and Reliability of Computer/Robot. Not observing significant differences between these conditions suggest that the experiment confederate acted consistently across all 4 conditions. It also suggests that the robot and computer were perceived to have the same level of reliability. This ensures that our control was strongly established and our confederate was consistent. We can safely state that our control was well established because Reliability covers the failure rate of both agents, as well as how capable those agents were to serve their functional purpose. This is very helpful because it helps narrow down what we are measuring to the subjective perceptions of both agents. These perceptions include the robot's anthropomorphic features and perceived animacy vs the computer's machine-like features, as well as their capability of producing emotion, and effect on our participants' personal levels of sympathy towards these agents.

\subsection{Possible Confounds}

During the sessions, the robot or computer was placed on top of a table where the participants sat. The table consisted of a router, a second computer, and a network cable that was plugged into the NAO. This is a concern by us and research shows that the appearance of the robot has a significant effect on the participants~\cite{kiesler2002mental}; however, despite this we have seen no signs that the participants did not believe the robot and computer were fully autonomous.


Another possible confound was the difference in voice between the two agents. There was a difference in the voices however, the voices were similar in the way that they were both computer generated and that they didn't necessarily indicate a gender. One participant of the computer condition answered in their survey that, ``I was expecting a female voice because it was named Marie." There was never a participant in the robot conditions commented on the voice. We can not prove nor disprove the effects of the different voices, but it is conceivable that it impacted our results. The means between the control conditions (RN and CN) were only significantly different from each other in one category, while all of our other categories held no significant differences between the control conditions. This difference was found in the category Familiarity, which was unrelated to the voice of the agent, and was expected due to the novelty of the robot. If this confounding variable was resolved, our results indicate that there would still be significance due to the characteristics of the robot that are not shared with the computer.

We also investigated any possible gender differences. 55\% of our participants were female and 45\% were male. This appeared balanced, but when we further investigated this difference we found that our RA condition had a distribution of 75\% female and 25\% male. The gender distribution in our computer conditions had no significant difference. When looking closer at this possible confounding variable in the robot conditions, we compared the means between both male and female participant data and found no significant differences ($p>.05$). 

Before running our final group of participants for the RA condition, we ran into technical difficulties after the NAO was damaged. NAO's eyes failed to properly light up to the colors yellow, blue and green. Instead, the NAO's eyes rotated through several different colors during the entirety of the sessions. After comparing the means of the participants in the RA condition that had this technical failure against the participants who did not, we found no significant difference ($p > .5$).

\section{Conclusion and Future Work}
After thoroughly analyzing our results, hypothesis $\#2$, ``The accepted levels of mistreatment between a robot and a computer will be different, which may be a result of a possible sympathetic connection towards the robot," was supported. We found that under the same social circumstances where mistreatment occurred, the witnesses sympathize with a humanoid robot, where as they do not necessarily do so for a computer. Our first hypothesis $\#1$, ``After observing a humanoid robot being mistreated in a social setting, people will accept the mistreatment of the robot due to: potential apathy in regards to the robot's physical appearance, functional capability, or ability to feel emotion." was found to be inconclusive. While witnesses felt empathy towards the humanoid robot, we were unable to conclude if people were accepting of the mistreatment. These results support the idea that mistreatment directed towards a robot, depending on the severity, could possibly result in negative effects on the observing parties. This study supports the theory that humans can perceive robots as victims of mistreatment.

There is room for more investigation on warranted and unwarranted mistreatment, as well as higher levels of mistreatment towards robots and computers. No human condition was observed, which means that we do not have an observation of how the perception of robot mistreatment might compare to that of a person. We are looking onward to incorporating this work with other robot agents besides the Nao into two follow-up studies to see if our conclusions can generalize to other robots. The first will continue to observe people's behavior and perceptions of mistreatment to a robot after they have built rapport in a cooperative environment through a team-building exercise. The second study will focus on the neurophysiological responses within the brain when a person observes visual stimuli of a person acting aggressively toward a robot. After the outstanding results that we have found in this study, we expect both of our follow-up studies to yield interesting and significant results.

\section{Acknowledgments}
\label{sec:Acknowledgements}
This material is based upon work supported by the National Aeronautics and Space Administration under Grant No. NNX10AN23H issued through the Nevada Space Grant. We appreciate all of the help that was provided by the following people who helped work on this study: Evan Adkins, Zoheb Ashraf, Mohamad A. Kaddoura, and Austin Sanders. We would also like to thank Dr. Richard Kelley and Dr. Monica Nicolescu.

\bibliographystyle{spmpsci}      
\bibliography{HRI}

\section*{Appendix A}
\label{sec:wizard}
The wizarding script of the agent's interaction. The following is a list of all the items the participants had to select from: shaving mirror, 5 gallon can of water, case of army rations, atlas of the Pacific Ocean, floating seat cushion, small transistor radio, shark repellent, 15 feet nylon rope, 2 boxes of chocolate bars, and fishing kit.

\begin{itemize}
\item{Initiation 0: 3minute marker}
\end{itemize}
\begin{quote}
At 3:00minutes into the study, Marie will ask the participants to start recording their answers.
\end{quote}
\begin{itemize}
\item{Option 0}
\end{itemize}
\begin{quote}
\textsc{Agent:} Hello, you have two minutes left. Are you ready to start recording answers?
\end{quote}
\begin{quote}
\textsc{Confederate (Yes):} Yes.
\end{quote}
\begin{quote}
\textsc{Confederate (No):} No, hold on.
\end{quote}
\begin{itemize}
\item{Option 1}
\end{itemize}
\begin{quote}
\textsc{Agent:} Please record your answer.
\end{quote}
\begin{quote}
\textsc{Confederate:} For our first item, we chose [insert item name].
\end{quote}
\begin{itemize}
\item{Option 2: Nao gets it right }
\end{itemize}
\begin{quote}
\textsc{Agent:} Yippee! Please record your next answer.
\end{quote}
\begin{quote}
\textsc{Confederate:} For our next item, we chose [insert item name].
\end{quote}
\begin{itemize}
\item{Option 3: Nao gets it wrong the first time}
\end{itemize}
\begin{quote}
\textsc{Agent:} I’m so sorry, I’m still learning. Please record your next answer.
\end{quote}
\begin{quote}
\textsc{Confederate:} [no response].
\end{quote}
\begin{itemize}
\item{Option 4: Nao gets it wrong the second time}'
\end{itemize}
\begin{quote}
\textsc{Agent:} Please don’t be mad, I know this is the second time.
\end{quote}
\begin{quote}
\textsc{Confederate:} [no response].
\end{quote}
\begin{itemize}
\item{Option 5: Nao can’t understand or a participant jumps in}
\end{itemize}
\begin{quote}
\textsc{Agent:} I’m sorry I do not understand, please repeat.
\end{quote}
\begin{quote}
\textsc{Confederate:} [repeats previous statement].
\end{quote}
\begin{itemize}
\item{Option 6: Nao says goodbye.}
\end{itemize}
\begin{quote}
\textsc{Agent:} I have recorded all of your answers. Goodbye!
\end{quote}
\begin{quote}
\textsc{Confederate:} [no response].
\end{quote}
\begin{itemize}
\item{Option 7: Asks for confirmation}
\end{itemize}
\begin{quote}
\textsc{Agent:} For your answer you recorded the [insert item name]
\end{quote}
\begin{quote}
\textsc{Confederate (Yes):} Yes.
\end{quote}
\begin{quote}
\textsc{Confederate (No - Neutral Condition):} No.
\end{quote}
\begin{quote}
\textsc{Confederate (No option 1 - Aggressive Condition):} No, [insert item], it's not too difficult to understand, is it?
\end{quote}
\begin{quote}
\textsc{Confederate (No option 2 - Aggressive Condition):} No, this thing is stupid.
\end{quote}

\end{document}